# Vehicle Detection in Deep Learning

Yao Xiao

Thesis submitted to the faculty of the Virginia Polytechnic Institute and State University
in partial fulfillment of the requirements for the degree of

Master of Science
In
Computer Engineering

A. Lynn Abbott
Daniel Pillis
R. Michael Buehrer

(May 3th, 2019)
Blacksburg, Virginia

Keywords: Vehicle Detection, Deep Learning, Convolutional Neural Networks, Image Processing, Architecture Design

# Vehicle Detection in Deep Learning

Yao Xiao

ABSTRACT

Computer vision is developing rapidly with the support of deep learning techniques. This thesis proposes an advanced vehicle-detection model based on an improvement to classical convolutional neural networks. The advanced model was applied against a vehicle detection benchmark and was built to detect on-road objects. First, we propose a high-level architecture for our advanced model, which utilizes different state-of-the-art deep learning techniques. Then, we utilize the residual neural networks and region proposal network to achieve competitive performance according to the vehicle detection benchmark. Lastly, we describe the developing trend of vehicle detection techniques and the future direction of research.

# Vehicle Detection in Deep Learning

Yao Xiao

GENERAL AUDIENCE ABSTRACT

Computer vision techniques are becoming increasingly popular. For example, face recognition is used to help police find criminals, vehicle detection is used to prevent drivers from serious traffic accidents, and written word recognition is used to convert written words into printed words. With the rapid development of vehicle detection given the use of deep learning techniques, there are still concerns about the performance of state-of-the-art vehicle detection techniques. For example, state-of-the-art vehicle detectors are restricted by the large variation of scales. People working on vehicle detection are developing techniques to solve this problem. This thesis proposes an advanced vehicle detection model, adopting one of the classical neural networks, which are the residual neural network and the region proposal network. The model utilizes the residual neural network as a feature extractor and the region proposal network to detect the potential objects' information.

# Dedication

*To my family*

# Acknowledgments

I would like to thank my thesis advisor, Professor Abbott. The door to Professor Abbott's office was always open for me to share any positive experiment results in my research. He continuously helped and provided guidance for my research, always helping me move in the right direction. I could not have completed this thesis without his help.

Professor Pillis is not only my advisor but a friend who has given me many helpful suggestions about future research and my life direction. Anyone would be lucky to work with him, and I am grateful for his help and suggestions.

I also want to thank Professor Buehrer who holds much knowledge on information theory and has helped me further understand the concept of mathematical knowledge in deep learning.

Last but not least, I want to express gratitude to my family and friends, who have supported me through all the obstacles I have encountered.

# Contents

# List of Figures

# List of Tables

# 1
# Introduction

This section introduces the rapid development of deep learning techniques that applies to the field of object recognition. We discuss some state-of-the-art convolutional neural networks, which have performed well in object detection and vehicle detection benchmarks. We briefly describe our advanced model.

## 1.1 Background

In 2012, Krizhevsky proposed AlexNet [5], which beat all the competitors in that year's ImageNet Large Scale Visual Recognition Challenge. AlexNet marked the beginning of the development of deep learning for visual applications, and since then, more neural network architectures have been proposed. These architectures include Vgg [6], ResNet [7], and Inception [8]. Deep learning has significantly improved the state-of-the-art performance in many fields, including natural language processing, computer vision, and recommender systems.

In computer vision, techniques take images or videos as input, process with fine-tuned algorithms and produce useful information for humans. Object detection is a major task in computer vision, and it can utilize the power of convolutional neural networks. For this research, we utilize several neural network architectures to detect vehicles.

This thesis focuses on the sub-area of vehicle detection. Drivers sometimes feel tired, exhausted and distracted while they are driving. Vehicle detection automatically detects vehicles, enabling drivers to make fast and sufficient decisions.

## 1.2 Motivation

Vehicle detection methods have been in development for several years in academia and industry. So far, some state-of-the-art object detection methods cannot achieve competitive performance on vehicle detection benchmarks. The main problems for vehicle detection are large variation of light, dense occlusion, and large variation of object scales.

Faster R-CNN [1] was proposed by He et al. for object detection in 2015. Faster R-CNN was the first to incorporate Region Proposal Network (RPN) as a Region of Interest (RoI) candidate extractor. Faster R-CNN shows great performance for the Pascal VOC [37] and COCO [38] 2D object detection benchmarks. But it has not performed well and only achieves 56.39% mean average precision on the KITTI [2] vehicle detection benchmark.

The un-competitive results of Faster R-CNN on the KITTI vehicle detection benchmark can be explained using one main reason – the large variation of vehicle scales. The RPN takes convolutional feature maps as input, and outputs potential RoI. The large variation of vehicle scales causes RPN to ignore small objects. We address the problem of large variation of scales in vehicle detection by proposing an extended RPN. We illustrate the comparison of images from original RPN and extended RPN in Figure 1.1.

We propose our whole architecture in Figure 1.2. We describe our model as three parts: Feature Extractor, Region Proposal Network, and Prediction Stage. The first stage is feature extractor, which takes raw image data as input and outputs feature maps. Our research contribution in feature extraction is that we incorporate an advanced ResNet which replaces residual mapping with identity mapping [9] to improve the performance. The second stage is a region proposal network, which takes feature maps as input and outputs RoI. Our research contribution in region proposal network is that we propose an extended RPN [1] which makes Faster R-CNN [1] more robust to handle the large variation of vehicle scales. The third stage is the prediction stage, which takes feature maps and RoIs as inputs, and makes final predictions.

## 1.3 Methodology

For preliminary work, we reviewed several papers related to object detection in deep learning. Our first step was to conduct experiments using Faster R-CNN on different object detection datasets, such as the Oxford-Pet [39] and the Pascal VOC [37] dataset. During the procedure, we became familiar with Tensorflow and Keras, two deep learning frameworks implemented by C++ and used by Python. Our goal is to improve the performance of Faster R-CNN in vehicle detection using the KITTI dataset [2].

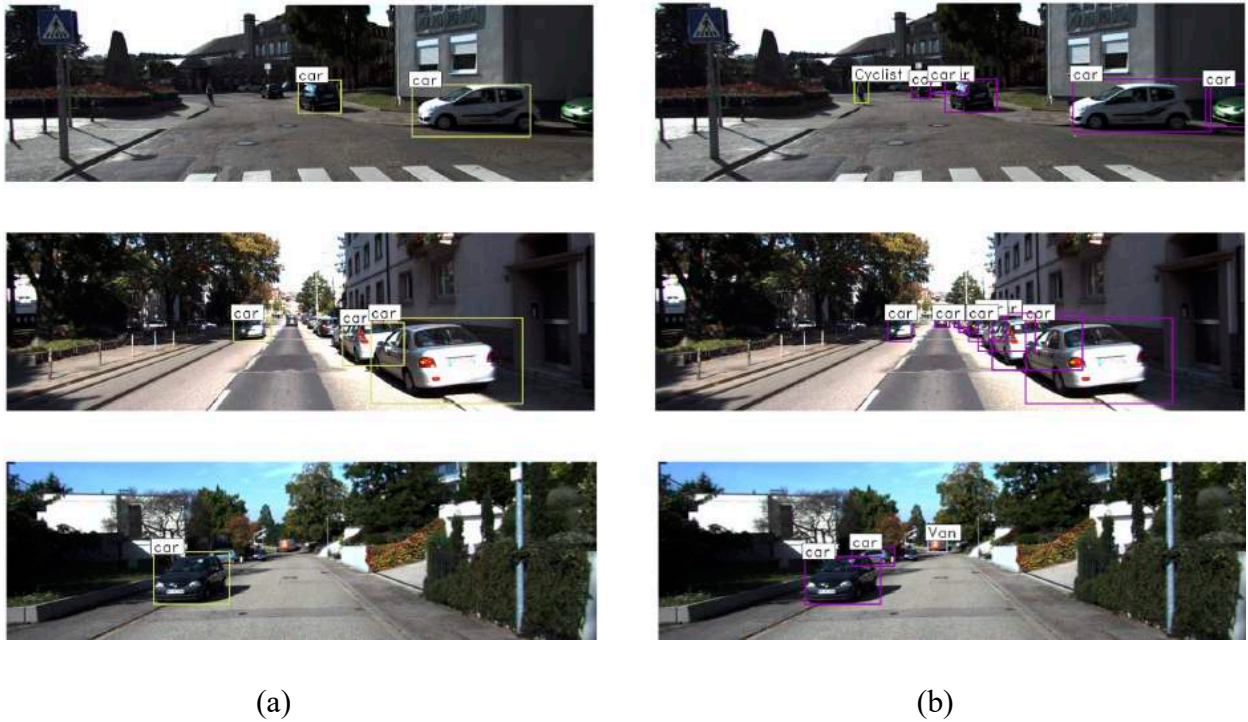

**Figure 1.1**: Examples for comparison. (a) Images from original models, (b) images from our proposed model.

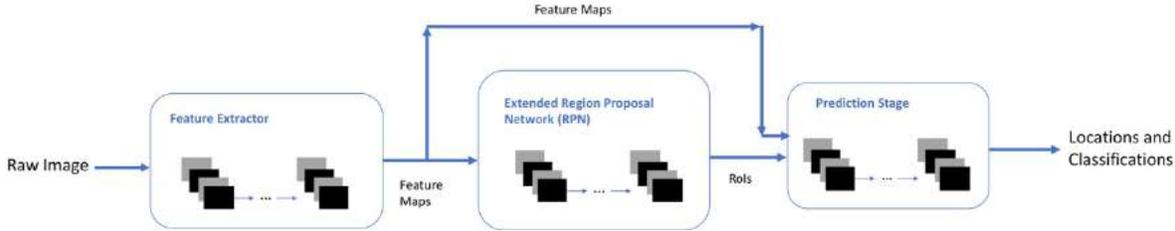

**Figure 1.2**: The illustration of the whole architecture of our advanced model. The first stage is feature extractor, which takes raw image data as input and outputs feature maps. The second stage is region proposal network, which takes feature maps as input and outputs with RoI. The third stage is prediction stage, which takes feature maps and RoI as inputs, and makes final predictions.

## 1.4 Related Work

In this section, we'll discuss four parts of related work: traditional vehicle detection methods, CNN vehicle detection methods, specialized vehicle detection methods, and KITTI 2D vehicle detection methods.

### 1.4.1 Traditional Vehicle Detection Methods

From the beginning of vehicle detection, researchers have proposed several traditional vehicle detection methods. Handcrafted features are used to determine the performance of methods. Histogram of Oriented Gradient (HOG) [14] and Haar-like [43] features are the most common-used features. One of the earliest real-time detectors is a cascaded detector [44], which achieves competitive accuracy. Deformable part-based models (DPM) [45] and Support Vector Machines (SVM) [15] are two well-known models of the part-based model approach.

### 1.4.2 CNN Vehicle Detection Methods

Several methods in deep learning have been proposed for vehicle detection. We discuss two directions of CNN methods: single-stage and two-stage.

#### 1.4.2.1 Two-stage

The first method is called region-based convolutional neural network (R-CNN) [13], which achieved good performance in vehicle detection. Region-based convolution neural network combines region proposal network with CNN, achieving better performance compared to HOG [14] features with a SVM classifier. A region-based convolutional neural network takes raw image data as an input and extracts region proposals. Then, the region proposals are fed into a CNN to extract features, and a support vector machine [15] is used to make predictions. The primitive R-CNN achieved 53% mean average precision in the Pascal VOC 2010 competition. Spatial Pyramid Pooling (SPP) [16] applies a convolution layer to the entire image and extracts features using SPP-net, which addresses the expensive computational cost of R-CNN.

#### 1.4.2.2 One-stage

For two-stage vehicle detector, the first stage is to extract features from raw image data and second stage is to make final predictions. In order to improve the vehicle detection speed, single-stage methods have been proposed. Single-shot detector (SSD) [4] is one of the state-of-the-art single-stage detectors, which make predictions by utilizing different resolutions of feature maps. You Only Look Once (YOLO) [3] is another type of single-stage detector, which makes predictions by regarding raw image data as a $7 \times 7$ grid.

### 1.4.3 Specialized Vehicle Detection Methods

The researchers address three main practical problems of vehicle detection: large variation of light, dense occlusion, and large variation of scales. We list some related work to handle the three problems.

### 1.4.3.1 Large Variation of Light

Saini [40] proposed a robust CNN model to handle the traffic light detection for autonomous vehicles, to address the problem of large variation of light. The model takes raw image data as input, extracts candidate region, and performs final traffic light detection and recognition. This CNN model improves the accuracy of traffic light detection from 97.73% to 99.78% with respect to SVM classifiers.

### 1.4.3.2 Dense Occlusion

In vehicle detection, dense occlusion makes it difficult to discriminate occluded vehicles. Phan [41] proposed their method to handle the dense occlusion from static surveillance cameras. The method consists of background subtraction, vehicle detection, and occlusion detection, which is to extract the occluded vehicles individually based on the external properties. This model improves the total accuracy of dense occluded vehicle detection from 61.19% to 84.09% compared to Ha's method [42].

### 1.4.3.3 Large Variation of Scales

Lu [35] recently proposed an improved Region Proposal Network (RPN) known as a scale-aware RPN, to address the problem of detecting vehicles at different scales. The scale-aware RPN consists of two sub-networks: one detects large proposals and another detects small proposals, and then feeds proposals into two separate XGBoost [36] classifiers to make final predictions.

### 1.4.4 KITTI 2D Vehicle Detection Methods

The KITTI is the most commonly used dataset in vehicle detection. Recently several papers have been proposed to improve the performance of vehicle detector on the KITTI dataset. Hu [46] proposed SINet to handle the problem of scale-sensitivity in the vehicle detection task. SINet incorporated a context-aware RoI pooling method and a multi-branch final prediction network. It achieved state-of-the-art performance in speed which is 37 FPS. Wei [47] proposed an enhanced CNN vehicle detection model to address the problem of large variation of vehicle scales. It proposed a deeper and wider neural network to extract features from image to achieve good prediction performance. Liu [48] proposed a multi-path processing technique to better preserve the original features information. It outperformed original Faster R-CNN by 16.5% mAP on the KITTI dataset. These works either increase the testing speed and accuracy of vehicle detector or succeed in handling the problem of large variation of vehicle scales. Our thesis's goal is to handle the problem of large variation of scales and increase the performance of vehicle detector.

# 2

# Theory

This chapter introduces technical concepts in deep learning and vehicle detection. We derive our vehicle detector by introducing several preliminary models and architectures.

## 2.1 Human Brain and Artificial Neural Networks

The biological neuron [17] is a special cell that receives and sends electrical signals and also processes electrical information. A typical neuron has four different functional parts: (1) dendrites, (2) an axon, (3) presynaptic terminals, and (4) soma [18]. The soma is the cell body, which provides nutrition and contains necessary substances such as DNA. The soma serves as the hub for conveying electrical information to dendrites and axons. The dendrites are the main parts that receive electrical information from other neurons. The axon conveys electrical information to other neurons. The end of the axon is divided into several branches, which communicate with other neurons at specific zones called synapses. The presynaptic terminal is the cell-transmitting signal, and the postsynaptic terminal is the cell-receiving signal. In the human brain, neurons are connected with each other through their synapses, which results in a biological neural network [19].

An artificial neural network [20] is a computer system that is inspired by the biological neural network. An artificial neural network is constructed by a collection of connected artificial neurons; each connection transits signals from one neuron to another. For a classical artificial neural network, the signal at the edge between artificial neurons is a real number, and the output of each artificial neuron is computed based on a non-linear function of the sum of the neuron's inputs. Artificial neurons typically have weights that adjust as learning proceeds. The weight increases or decreases the strength of the signal. Artificial neurons have a threshold that marks when a signal is sent for a particular input, and artificial neurons are aggregated into layers. Different layers perform different kinds of transformations on their inputs. Figure 2.1 displays the biological neuron and the artificial neuron.

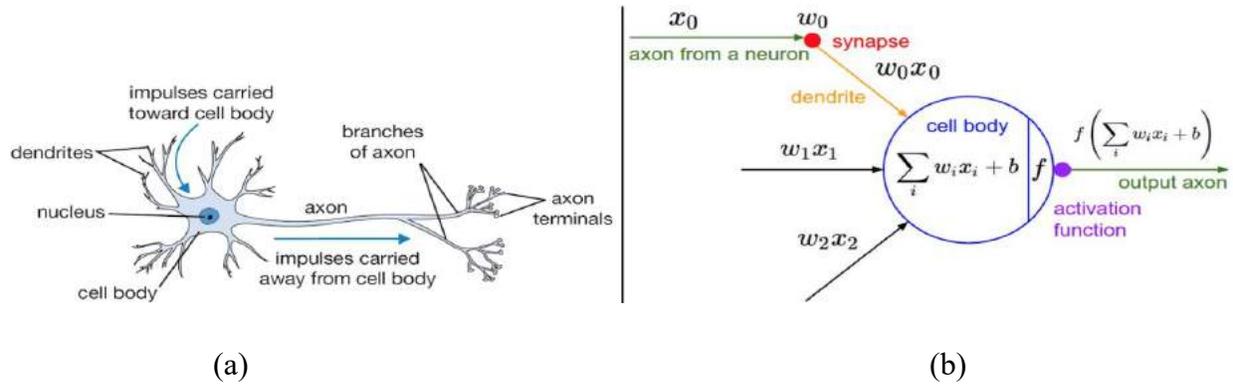

(a)            (b)

**Figure 2.1**: (a) Architecture of an biological neuron, (b) architecture of an artificial neuron (figure taken from [10]).

## 2.2 Feedforward Neural Network

The feedforward neural network is the most common model in artificial neural networks. The feedforward neural network propagates signals in one direction, from input nodes to output nodes [21].

Feedforward neural networks comprise three main properties:

(1) Neurons are aggregated into layers, with the first layer taking in inputs and the last layer producing outputs. The middle layers are called hidden layers because they are connected to the previous layer and the next layer.
(2) In the given layer, the output of each neuron is connected to neurons in the next layer.
(3) In the given layer, each neuron has no connection with other neurons in the same layer.

We illustrate a simple model of the feedforward neural network in Figure 2.2.

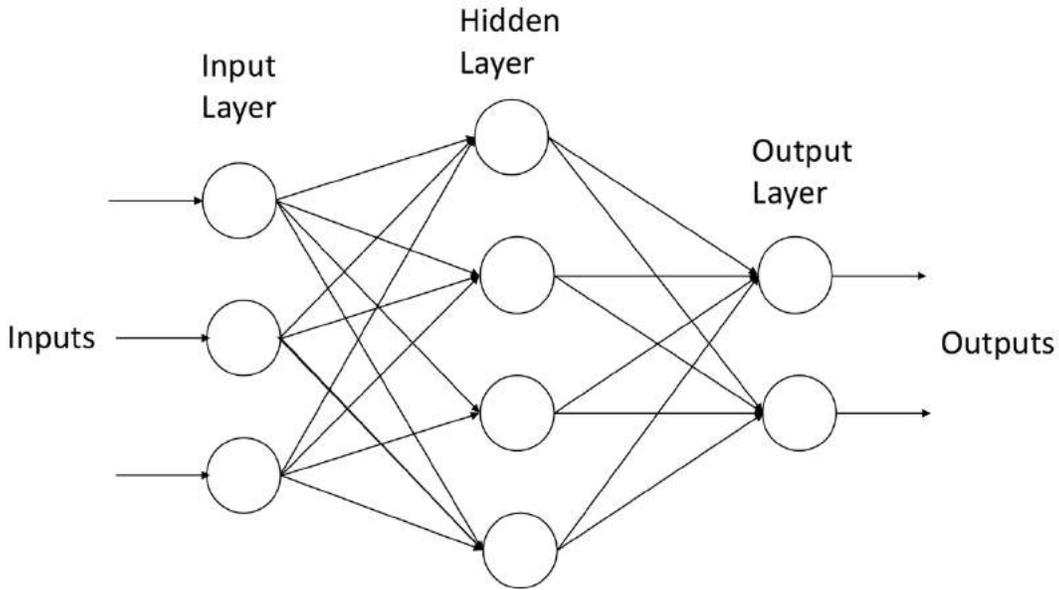

**Figure 2.2**: A simple model of the feedforward neural network.

### 2.2.1 Activation Functions

One of the essential factors that influences the capability of artificial neural networks to approximate non-convex functions is non-linear activation functions. An activation function takes a scalar value as an input and applies transformation. We review several activation functions and properties as below:

**(1) Sigmoid**  The sigmoid activation function is defined as below:

$$y = f(x) = \frac{1}{1+e^{-x}} \qquad (2.3)$$

It takes a real value $x$ as an input and converts into a value between 0 and 1.

**(2) Hyperbolic Tangent**  The hyperbolic tangent activation function is defined as below:

$$y = f(x) = \frac{e^x - e^{-x}}{e^x + e^{-x}} \qquad (2.4)$$

It takes a real value as an input and converts into a value between -1 and 1.

**(3) Rectified Linear Unit**  The ReLU activation function is defined as below:

$$y = f(x) = \max(0, x) \qquad (2.5)$$

The ReLU is the most commonly used activation function in deep neural networks because it has exhibited better performance for converging, compared to sigmoid and hyperbolic tangent activation functions. Essentially, ReLU ignores all of the negative inputs.

These three types of activation functions are illustrated in Figure 2.3.

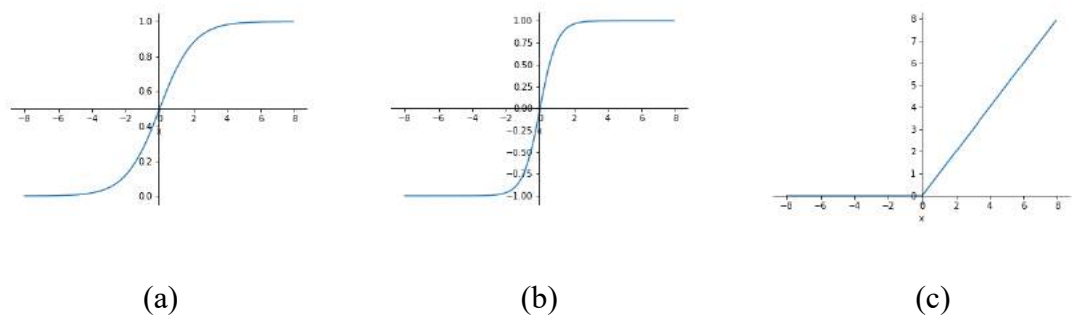

(a)           (b)           (c)

**Figure 2.3**: Illustration of three activation functions. (a) sigmoid, (b) tanh, and (c) ReLU.

## 2.3 Convolutional Neural Network

Convolutional Neural Network is a special type of feedforward neural networks and are designed to operate on images. A convolutional neural network consists of an input and output layer, as well as several hidden layers including the convolutional layer, the activation layer, the polling layer, the fully connected layer, and the normalization layer. We review all of the distinct layers as below.

### 2.3.1 Convolutional Layer

If we consider that the input of a convolutional layer is a raw image with size $32 \times 32 \times 3$ ($width \times height \times channels$), we assume our convolutional layer is a "filter" of size $5 \times 5 \times 3$ ($width \times height \times channels$). For forward passing, we "slide" (convolve) each filter across the input image data and compute the sum of the dot products between the entries of the filter and the input. Thus, we have a two-dimensional feature map for a single filter.

When CNN deals with high-dimensional data like images, each neuron of the same convolutional layer is connected to a small portion of the inputs, which are called the *receptive field*. Another important property is *spatial arrangement,* which reduces the dimensions of inputs. As we mentioned before, a convolutional layer takes an image of size $32 \times 32 \times 3$ as an input and produces the output of a feature map of size $28 \times 28 \times 6$. However, we applied two strategies to avoid and control this effect:

(1) The *depth* controls the number of the neurons in a layer connected to the same region of the inputs.
(2) *Stride* is the pixel number that the amount of steps filter moves. For example, if stride is 1, then we move filters one pixel each time. If we set a higher stride length, the receptive field overlaps less and result in smaller size outputs.
(3) *Zero-padding* is a key method that keeps the output dimension the same as the input dimension.

### 2.3.2 $1 \times 1$ Convolution

$1 \times 1$ convolution was firstly proposed by Network in Network [25]. The goal of $1 \times 1$ convolution is to modify the dimension of inputs without losing information. Since we have an input of size $H \times W \times D$ ($H$ denotes height, $W$ denotes width, and $D$ denotes depth) and a convolutional layer of size $1 \times 1 \times D'$, then we can obtain an output of size $H \times W \times D'$, which converts the depth of input from $D$ to $D'$.

### 2.3.3 Pooling Layer

The pooling layer is another important layer in CNN, which is a form of non-linear down-sampling. Max pooling is the most commonly used pooling operation, which partitions the inputs into several separated rectangles and outputs a maximum value for each sub-region, illustrated in Figure 2.4. The pooling layer operates independently in every depth of the input and resizes input spatially. A common form is a pooling layer with filters of size $2 \times 2$ and stride length of 2 applied on the inputs, ignoring 75% of the parameters. Pooling layer was proposed based on the assumption that the feature's relative location to other features is more important than its exact location in the input image.

### 2.3.4 Activation Layer

The activation layer applies the activation function on the inputs to incorporate the nonlinearity of the network. The ReLU function is the most commonly used activation function. Figure 2.5 illustrates the ReLU activation layer applied to the inputs.

### 2.3.5 Fully-Connected Layer

The most expensive operation of computation in a neural network corresponds to fully-connected layers. Neurons in a given fully-connected layer are connected to every neuron in the previous layer. We have illustrated fully-connected layers in Figure 2.6.

## 2.3.6 Loss Layer

A loss layer is normally the last layer of the network. Its purpose is to compute the training penalties between the predictions and ground-truth labels. For different types of tasks, we have different loss functions. We apply softmax loss function on classification problems.

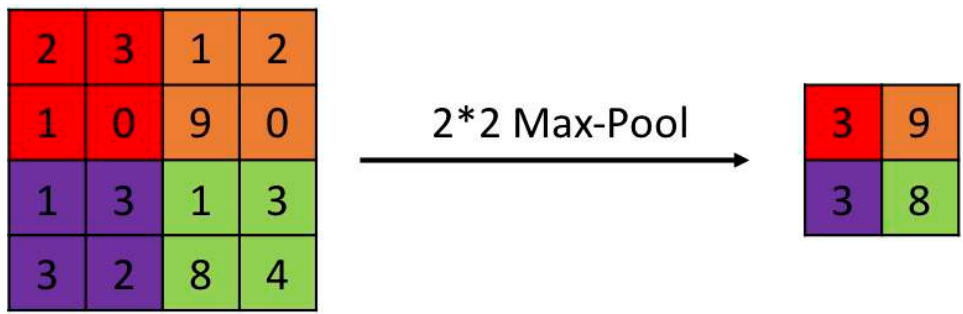

**Figure 2.4**: This figure demonstrates max pooling operations on the inputs. The value of same color region in the right rectangle corresponds to the maximum value of the same color region in the left rectangle.

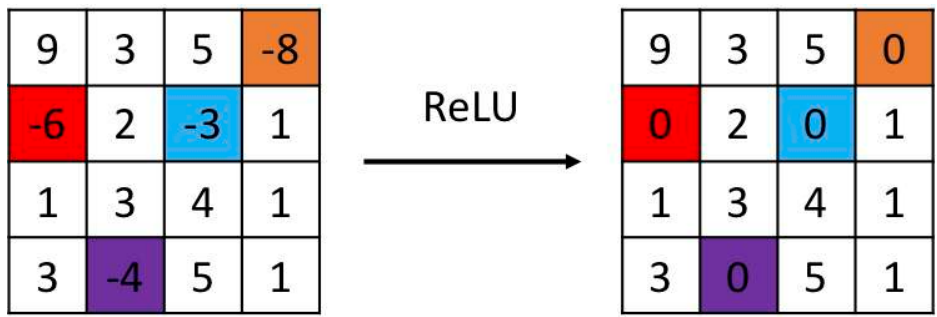

**Figure 2.5**: The illustration displays the ReLU activation layer applied to the inputs. ReLU converts negative values to zero.

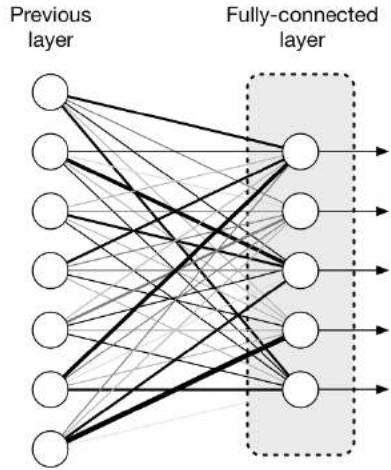

**Figure 2.6**: The illustration of a fully-connected layer. Neurons in the fully-connected layer are connected to every neuron in the previous layer (figure taken from [21]).

## 2.4 Training Techniques

### 2.4.1 Transfer Learning

It is extremely time-consuming to train a deep neural network for a complicated image recognition task. Thus, Pratt proposed an efficient method named transfer learning for storing knowledge gained from a low-level image recognition task [29]. For example, we trained a deep neural network on the ImageNet object detection dataset for recognizing vehicles, and we further train this deep neural network to recognize trucks and vans.

### 2.4.2 Optimization Methods

#### 2.4.2.1 Dropout

As we discussed before, fully-connected layers occupy most of the parameters in the network, which makes the network prone to overfitting. By utilizing dropout technique, during training, every node in the network is determined to be "dropped out" (concisely, ignored) with a probability $p$. Normally, we set $p$ to be 0.5 during training, and we average all of the nodes with probability $p$ during testing. Dropout avoids training all of the nodes in the network, and results in improving training speed and performance. We illustrate the dropout procedure in Figure 2.7.

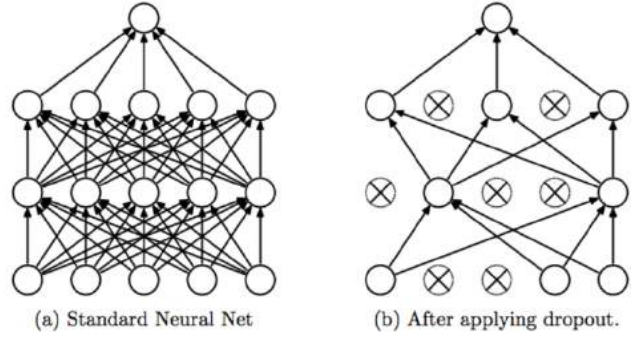

**Figure 2.7**: Illustration of dropout procedure (figure taken from [49]).

### 2.4.2.2 Batch Normalization

Training a deep neural network is complicated based on the fact that the input distributions for each layer change internally as the parameters of all the preceding layers change. We define the change of the distributions of internal neurons as *Internal Covariate Shift* [26]. The *internal covariate shift* makes a gradient of inputs gradually vanish but gets the network stuck in the saturated regime and makes training slower. We address *internal covariate shift* by making normalization on part of the network and applying normalization for every training mini-batch, which is called *Batch Normalization*.

Given the mini-batch inputs $\mathcal{B} = \{x_1, x_2, \ldots, x_n\}$ and outputs $\{y_1, y_2, \ldots, y_n\}$. We denote *batch normalization* to operations as follows:

$$BN_{\gamma,\beta}: \{x_1, x_2, \ldots, x_n\} \rightarrow \{y_1, y_2, \ldots, y_n\},$$

where $\gamma, \beta$ are learnable parameters during training. We propose BN algorithm as below:

**Input**: inputs of $x$ in one mini-batch: $\mathcal{B} = \{x_1, x_2, \ldots, x_n\}$;
    Learnable parameters: $\gamma, \beta$

**Output**: $\{y_1, y_2, \ldots, y_n\}$

$$\mu_\mathcal{B} \leftarrow \frac{1}{n}\sum_{i=1}^{n} x_i \qquad \text{// mini-batch mean value}$$

$$\sigma_\mathcal{B}^2 \leftarrow \frac{1}{n}\sum_{i=1}^{n}(x_i - \mu_\mathcal{B})^2 \qquad \text{// mini-batch variance value}$$

$$\hat{x}_i \leftarrow \frac{x_i - \mu_\mathcal{B}}{\sqrt{\sigma_\mathcal{B}^2 + \epsilon}} \qquad \text{// normalization operation}$$

$$y_i \leftarrow \gamma \hat{x}_i + \beta = BN_{\gamma,\beta}(x_i) \qquad \text{// scale and shift}$$

## 2.5 Classical Convolutional Architectures

With the development of deep learning, a number of classical convolutional architectures have been proposed since 1990. In this section, we introduce and describe the details of several convolutional architectures.

### 2.5.1 LeNet-5

In 1998, LeCun proposed the first primitive CNN called LeNet-5, which was used to deal with the recognition problem of zip codes and digits [27]. LeNet-5 introduced and used some basic ideas of CNN, which performed well for document recognition. However, there still exist some problems with respect to the state-of-the-art CNN architectures. Figure 2.8 displays the architecture of LeNet-5.

The research contributions of LeNet-5 are organized as follows:

i. Apply normalization operation to the inputs and utilize sparse matrix for connections between layers to improve training speed.
ii. Introduce a classical sequence for CNN: convolution → pooling → activation.
iii. Adopt average pooling as a pooling function, hyperbolic tangent and sigmoid function as activation functions, fully-connected layers as the classifier, and mean-square-error as the loss function.

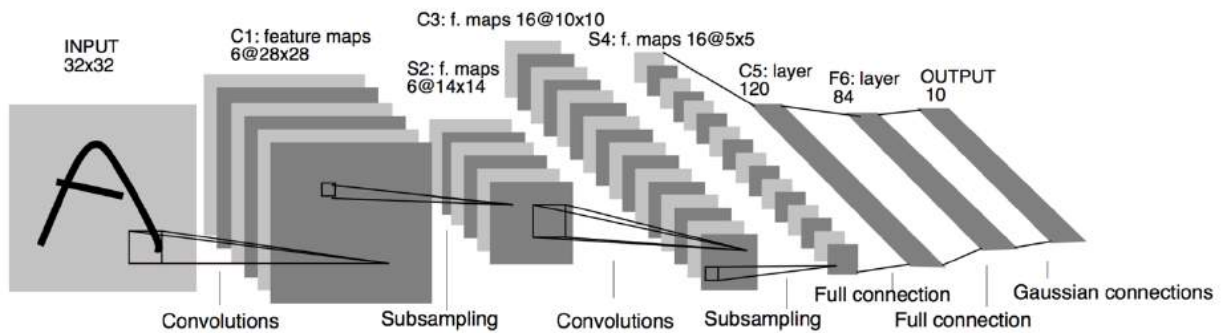

**Figure 2.8**: Illustration of LeNet-5, one of the original architectures for CNN (figure taken from [27]).

## 2.5.2 AlexNet

AlexNet is one of the most influential architectures in computer vision [28]. AlexNet achieved a top-5 error of 15.3% compared to the second runner up of 26.1% in the ImageNet Large Scale Visual Recognition Challenge of 2012. The most important contribution of AlexNet is that it utilizes GPU to solve the expensive computation of training deep neural networks. We illustrate the architecture of AlexNet in Figure 2.9. The properties of AlexNet are listed as follows:

  i. Consist of five convolutional layers, three max-pooling layers, and three fully-connected layers, which is deeper and has more parameters (60 million) than LeNet-5.
 ii. Adopt non-saturating ReLU activation function, which is proven to be better than the tanh and the sigmoid activation function.
iii. Utilize data augmentation for inputs, dropout layer, and stochastic gradient descent for preventing overfitting.

## 2.5.3 Vgg

Vgg [6] was proposed to further prove the idea that the performance of the CNN improves as the depth of networks increases. Vgg adopts multiple $3 \times 3$ filters to increase the depth of the network based on the fact that two $3 \times 3$ filters have the same receptive field as one $5 \times 5$ filter, and three $3 \times 3$ filters have the same receptive field as one $7 \times 7$ filter. Figure 2.10 illustrates the same receptive field. Meanwhile, three $3 \times 3$ filters achieve the same performance as one $7 \times 7$ filter but are approximately 3.7 times faster because the number of the parameters of three $3 \times 3$ filters is approximately half of the number of the parameters of one $7 \times 7$ filter.

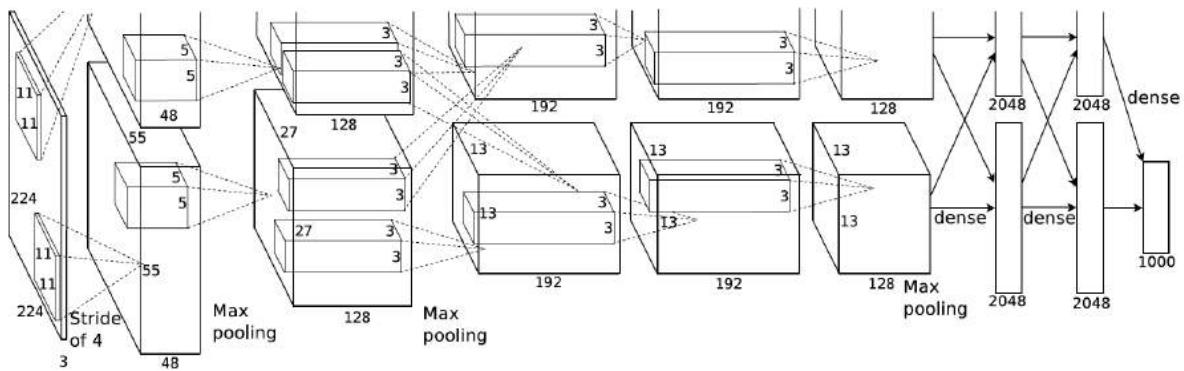

**Figure 2.9**: The illustration of the architecture of AlexNet. Two GPUs are separately utilized for the stacked layers of the top and bottom parts (figure taken from [6]).

## 2.5.4 ResNet

We use deeper CNN based on the assumption that deep neural networks achieve better performance in image recognition. However, deeper neural networks are found to be more difficult to train. He proposed the Deep Residual Network (ResNet) [7] to address the problem of the difficulty of training by learning residual functions with respect to the inputs. The idea behind residual learning is to incorporate a skip connection between multiple convolutional layers. Figure 2.11 illustrates the residual block. The learning output is $H(x)$, the input is $x$, and the original mapping of multiple convolutional layers is $F(x)$. Thus, we end up with $F(x) = H(x) - x$. When the network goes deeper, it is able to choose an identity shortcut if the network finds $F(x)$ is useless. Residual learning makes deep neural networks easier to train and achieves better performance than a plain network.

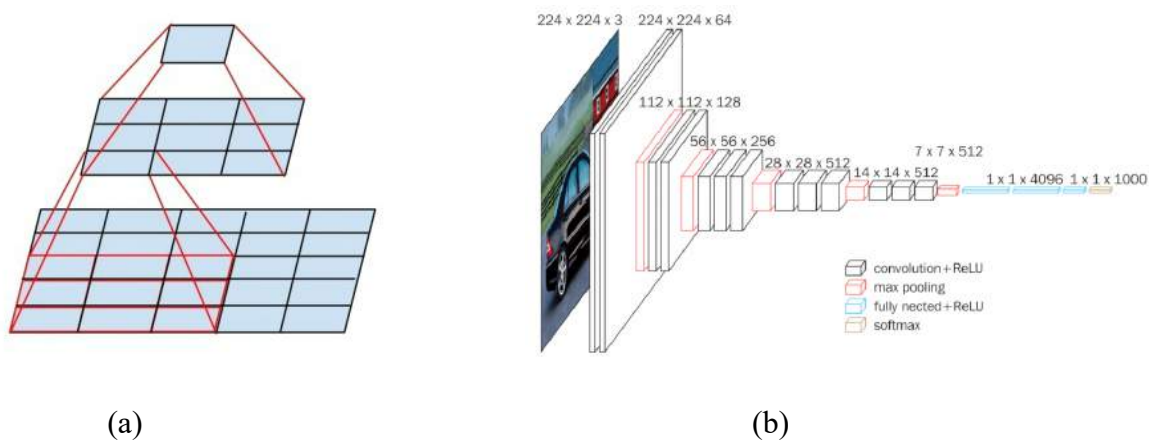

(a) (b)

**Figure 2.10**: (a) Two 3×3 filters have the same receptive field as one 5×5 filter, (b) illustration of the architecture of Vgg-16 (figures taken from [6]).

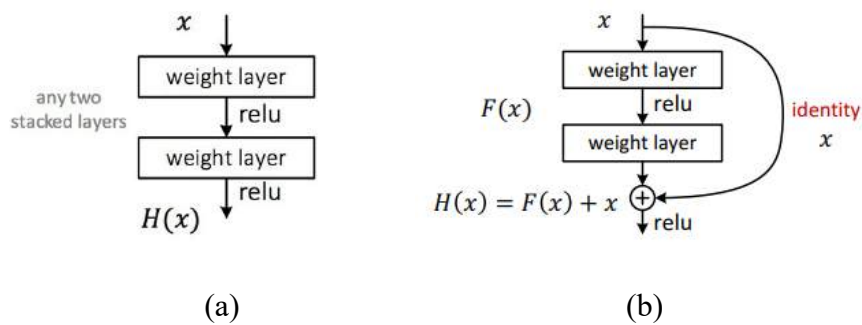

(a) (b)

**Figure 2.11**: (a) The architecture of plain networks, (b) the architecture of the residual block (figures taken from [7]).

# 3

# Methods

In this chapter, we describe our model as three parts: Feature Extractor, Region Proposal Network, and Prediction Stage [9]. We implement our vehicle detector on Tensorflow and Keras framework, which is available at https://github.com/PatrickXYS/Reproduce_frcnn. The reason we use Tensorflow is that the mechanism of Tensorflow makes it easy to deploy in multiple machines, which are heavily used in industry.

## 3.1 ImageNet Object Detection Pre-Training

As we mentioned before, we incorporate transfer learning in order to save training and testing time for vehicle detection. Thus, we implement three feature extractors: Vgg-16, ResNet-50, and ResNet-101. These three models were pre-trained on ImageNet object detection dataset [5] by He [7] and Simonyan [6].

### 3.1.1 ImageNet Object Detection Dataset

The ImageNet dataset [5] is a collection of natural images for investigating the performance of image recognition methods. The ImageNet dataset consists of 456,567 images for training, 20,121 images for validation, and 60,000 for testing. The dataset is divided into 1,000 categories that are made up of 20 main categories as follows: airplane, bicycle, bird, watercraft, wine bottle, bus, car, domestic cat, chair, cattle, dog, horse, motorcycle, person, sheep, sofa, table, flower pot, train, and TV or monitor. Each object in each image exists an annotation label for object category and object bounding box, which denotes the location of the object if it exists. Figure 3.1 illustrates the high-level classes of the ImageNet dataset.

## 3.1.2 Feature Extractor Architecture

We incorporate three types of feature extractors to deal with the ImageNet object detection task: Vgg-16, ResNet-50, and ResNet-101. Each of them is utilized in the feature extractor, RPN, and prediction stage, which will be discussed in Section 3.2. Figure 3.2 illustrates the architectures of Vgg-16 and ResNet-34. We use Table 1 to depict the difference between ResNet-34, ResNet-50, and ResNet-101; thus, we will not plot the whole architectures of ResNet-50 and ResNet-101.

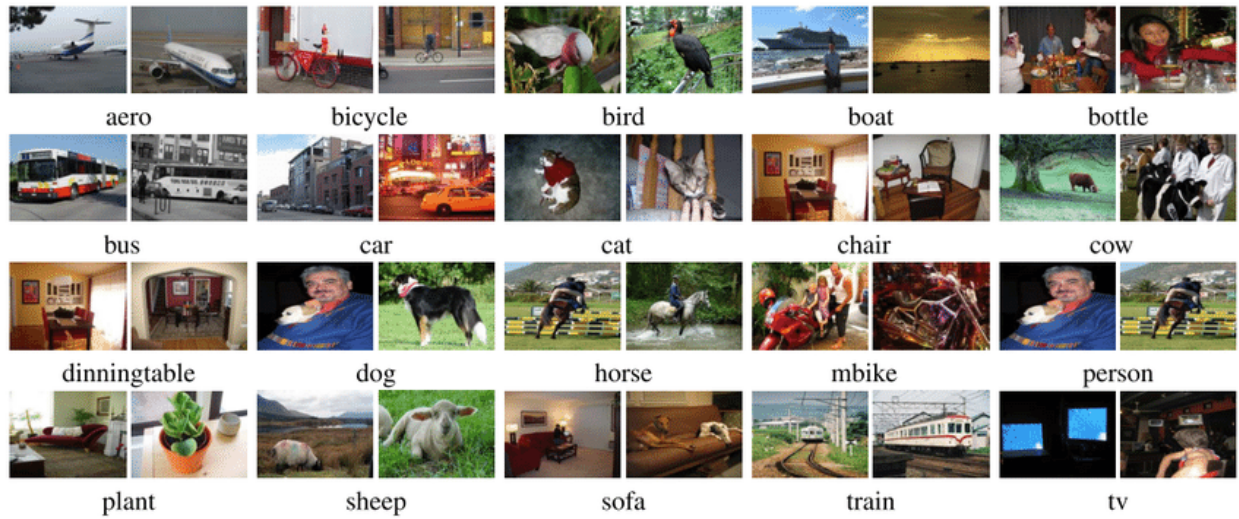

**Figure 3.1**: Twenty main categories of object in ImageNet dataset (figure taken from [5]).

**Table 1:** The whole architectures of ResNet-34, ResNet-50, and ResNet-101 are illustrated by separate blocks.

| Layer name | Output size | 34-layer | 50-layer | 101-layer |
|---|---|---|---|---|
| Conv_1 | $112 \times 112$ | $7 \times 7$, 64, Stride 2 | | |
| Conv_2 | $56 \times 56$ | $3 \times 3$ max pool, stride 2 | | |
| | | $\begin{pmatrix} 3 \times 3 & 64 \\ 3 \times 3 & 64 \end{pmatrix} \times 3$ | $\begin{pmatrix} 1 \times 1 & 64 \\ 3 \times 3 & 64 \\ 1 \times 1 & 256 \end{pmatrix} \times 3$ | $\begin{pmatrix} 1 \times 1 & 64 \\ 3 \times 3 & 64 \\ 1 \times 1 & 256 \end{pmatrix} \times 3$ |
| Conv_3 | $28 \times 28$ | $\begin{pmatrix} 3 \times 3 & 128 \\ 3 \times 3 & 128 \end{pmatrix} \times 4$ | $\begin{pmatrix} 1 \times 1 & 128 \\ 3 \times 3 & 128 \\ 1 \times 1 & 512 \end{pmatrix} \times 4$ | $\begin{pmatrix} 1 \times 1 & 128 \\ 3 \times 3 & 128 \\ 1 \times 1 & 512 \end{pmatrix} \times 4$ |
| Conv_4 | $14 \times 14$ | $\begin{pmatrix} 3 \times 3 & 256 \\ 3 \times 3 & 256 \end{pmatrix} \times 6$ | $\begin{pmatrix} 1 \times 1 & 256 \\ 3 \times 3 & 256 \\ 1 \times 1 & 1024 \end{pmatrix} \times 6$ | $\begin{pmatrix} 1 \times 1 & 256 \\ 3 \times 3 & 256 \\ 1 \times 1 & 1024 \end{pmatrix} \times 23$ |
| Conv5 | $7 \times 7$ | $\begin{pmatrix} 3 \times 3 & 512 \\ 3 \times 3 & 512 \end{pmatrix} \times 3$ | $\begin{pmatrix} 1 \times 1 & 512 \\ 3 \times 3 & 512 \\ 1 \times 1 & 2048 \end{pmatrix} \times 3$ | $\begin{pmatrix} 1 \times 1 & 512 \\ 3 \times 3 & 512 \\ 1 \times 1 & 2048 \end{pmatrix} \times 3$ |
| | $1 \times 1$ | Average pool, 1000-d fc, softmax | | |
| Flops | | $3.6 \times 10^9$ | $3.8 \times 10^9$ | $7.6 \times 10^9$ |

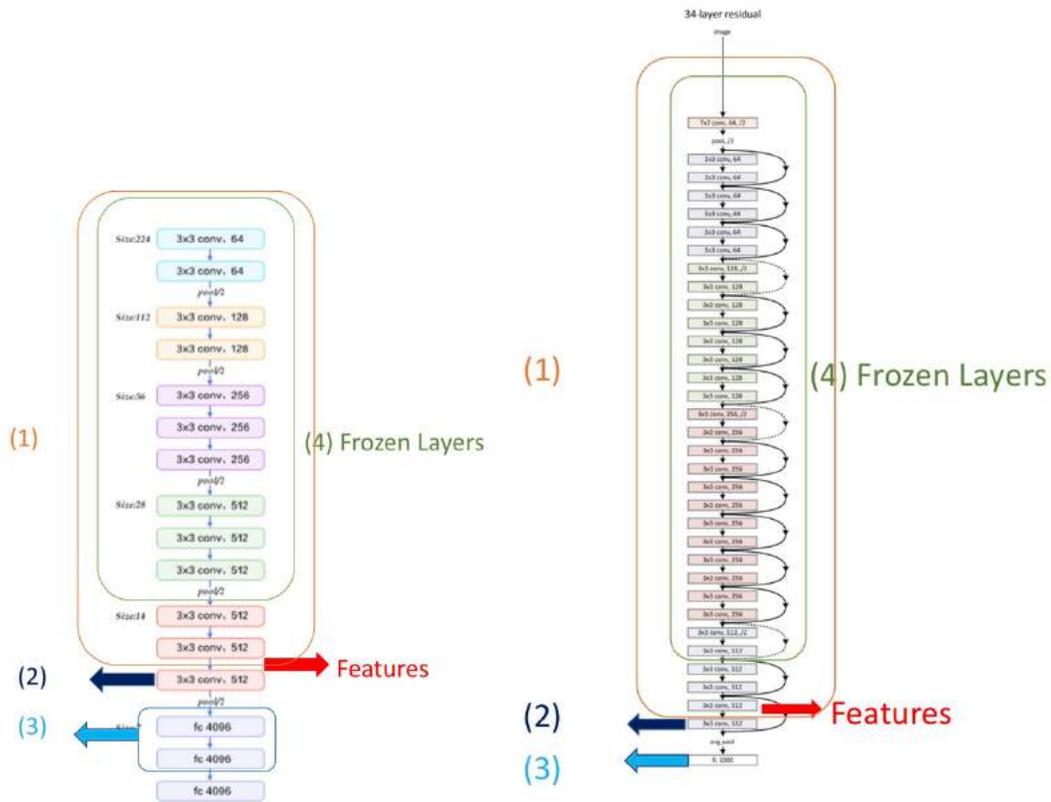

**Figure 3.2**: The entire architectures of Vgg-16 and ResNet-34. (1) Architecture of feature extractor. We trained the last two convolutional layers of Vgg-16, and the last three convolutional layers of ResNet-34. (2) The first convolutional layer of RPN. (3) Two fully-connected layers of Vgg-16 serve as layers of the prediction stage, and one fully-connected layer of ResNet-34 serves as the first fully-connected layer of the prediction stage. We trained all the layers in the feature extractor, RPN, and prediction stage except (4) frozen layers.

## 3.2 Vehicle Detection

The goal of this thesis is to build an advanced vehicle detector to recognize cars, pedestrians, cyclist, van, truck, tram, miscellaneous (e.g., trailers), and background based on the image dataset captured from the camera mounted on the front of the driving vehicles. We simultaneously predict the scores and coordinates of the objects, which are combined into detection. The following sections cover the KITTI dataset, model architecture, and implementation details.

### 3.2.1 KITTI Dataset

The KITTI vehicle detection benchmark [2] consists of 7,481 images. We split dataset into training and validation datasets consisting of 5,000 images and 1,481 images, respectively. We test our advanced Faster R-CNN model with 1,000 images. The size of each raw image data is $1242 \times 375 \times 3$. The raw images were recorded from a moving platform while driving in Germany [2].

The raw dataset can be evaluated at three levels of difficulty: easy, moderate, and hard. The easy level images consist of bounding boxes whose minimum value of the height is 40 pixels, the moderate level images consist of bounding boxes whose minimum value of height is 25 pixels, and the hard level images consist of bounding boxes whose minimum value of height is 25 pixels but with heavy occlusion. Thus, we take the moderate difficulty level as the evaluation level.

The corresponding bounding boxes of cars are annotated with categories and coordinates. The reference of the bounding box coordinates is the upper left corner point.

### 3.2.2 Model Architecture

We organize our model into three following sections: Feature extractor (Vgg-16, ResNet-50, ResNet-101, Advanced ResNet-101), Region Proposal Network (RPN), and Prediction Stage.

### 3.2.2.1 Feature Extractor

As we discussed in Section 2.5, we adopt Vgg-16, ResNet-50, and ResNet-101 as our feature extractors. The purpose of the feature extractor is to extract features from the raw image data. To save training time, we "freeze" the weights of most of the convolutional layers of a pre-trained model. Meanwhile, we extract the feature maps from the output of the last convolutional layers. The following fully connected layers are utilized for RPN and prediction stage, which will be discussed later. We illustrate the frozen layers of feature extractor in Figure 3.2 which takes Vgg-16 and ResNet-34 as examples. The frozen layers consist of most of the convolutional layers in Vgg-16 and ResNet-34. We propose the whole architecture in Figure 3.6.

Although the original ResNet has proven to perform well and faster in deep learning, we still believe there is a better way to perform residual mapping, also called identity mapping [9]. We illustrate the comparison between residual mapping and identity mapping in Figure 3.3. Identity mapping achieves better performance and faster training speed for image recognition. As we discussed before, original residual mapping are computed as follows:

$$y_k = F(x_k, W_k) + x_k \quad (3.1)$$

However, $y_k$ will not be directly considered an input of the next residual unit, which means:

$$x_{k+1} = f(y_k) \qquad (3.2)$$

where the $f$ function denotes the ReLU activation function, $x_k$ and $x_{k+1}$ denote the inputs of $k$ layer and $k+1$ layer. The identity mapping denotes that we apply a "clean" shortcut rather than applying a ReLU function. We illustrate the architecture of the original and advanced model in Figure 3.4. In our experiments, we replace our original ResNet-101 with an advanced ResNet-101 using identity mapping that have exhibited an improvement in performance.

### 3.2.2.2 Region Proposal Network (RPN)

Region Proposal Network [1] is a key point of Faster R-CNN, which replaces Selective-Search [31] and decreases generating time for region of interest from 2 seconds per image to 0.01 seconds per image. Region Proposal Network takes raw image data as input and produces region of interest, which denotes the possibility of object existence.

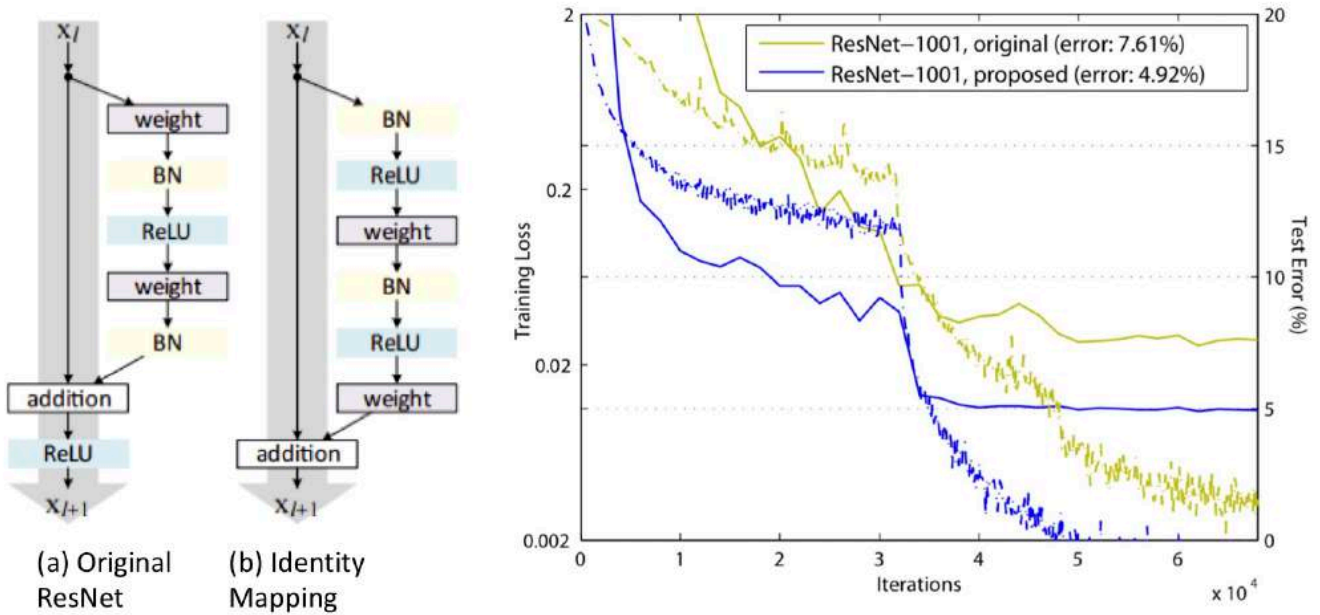

**Figure 3.3**: **Left**: (a) the original architecture, (b) the identity mapping (figures taken from [9]). The advanced architecture performs a "clean" mapping. **Right**: Ren trained original and advanced ResNet-1001 on CIFAR-10 [18]. Solid lines denote test errors, and dashed lines denote training errors.

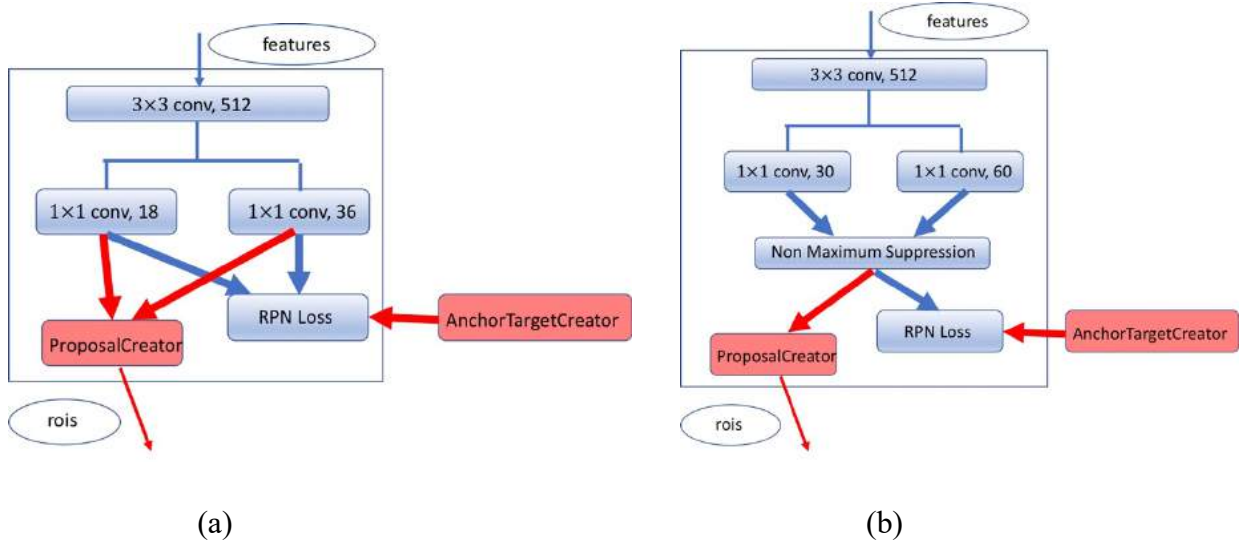

(a) (b)

**Figure 3.4**: Comparison between original RPN and extended RPN architectures. Original RPN uses anchor sizes of [128, 256, 512], and the extended RPN uses anchor sizes of [32, 64, 128, 256, 512]. (a) Illustration of the architecture of original RPN, (b) illustration of the architecture of extended RPN.

A Region Proposal Network consists of one $3 \times 3$ convolutional layer, two $1 \times 1$ convolutional layers, one non-maximum suppression layer, and one loss layer, which are shown in Figure 3.4. A Region Proposal Network applies a small $n \times n$ convolutional layer, which slides over the extracted feature map produced by the last convolutional layer of feature extractor. The idea behind this operation is that we apply $k$ sizes of "anchors" on the extracted feature maps, which are combined with three scale ratios: 128, 256, and 512 pixels, and three aspect ratios: 1:1, 1:2, and 2:1. The output of this $n \times n$ convolutional layer is converted into a 1024-D vector and then processed by two $1 \times 1$ convolutional layers: a bounding box regression layer and a bounding box classification layer, which produce bounding box coordinates and scores, respectively. To address the problem of large variation of vehicle scales, we propose the extended Region Proposal Network. The Extended RPN replaces original scale ratios [128, 256, 512] with extended scale ratios [32, 64, 128, 256, 512].

### 3.2.2.3 Prediction Stage

As we discussed before, the RPN takes raw image data and feature maps as inputs and produces regional proposals. The prediction stage randomly samples 128 region of interest from the region proposals. The region of interest pooling layer projects 128 different sizes of RoIs into the same $H \times W$ size vectors which makes it easy to connect to fully-connected layers. We then have two output layers for the probability of each object's category and the coordinates of the bounding box, respectively. We illustrate the whole architecture and high-level architecture of the prediction stage in Figure 3.5. We also show the whole architecture in Figure 3.6.

### 3.2.2.4 Sharing Computation between RPN and Prediction Stage

So far, we have three feature extractors, Vgg-16, ResNet-50, and ResNet-101, to produce convolutional feature maps. We also have a Region Proposal Network, which takes raw image data and convolutional feature maps as input and outputs RoIs, which denote the probability and coordinates of the bounding box. Furthermore, we have a prediction stage, which takes feature maps and RoIs as inputs and produces the object score and the coordinates of the bounding box.

The RPN and the prediction stage both take feature maps as input. Thus, we train both stages jointly with the feature extractor to build an end-to-end training pipeline. We illustrate our whole vehicle detection pipeline in Figure 3.6.

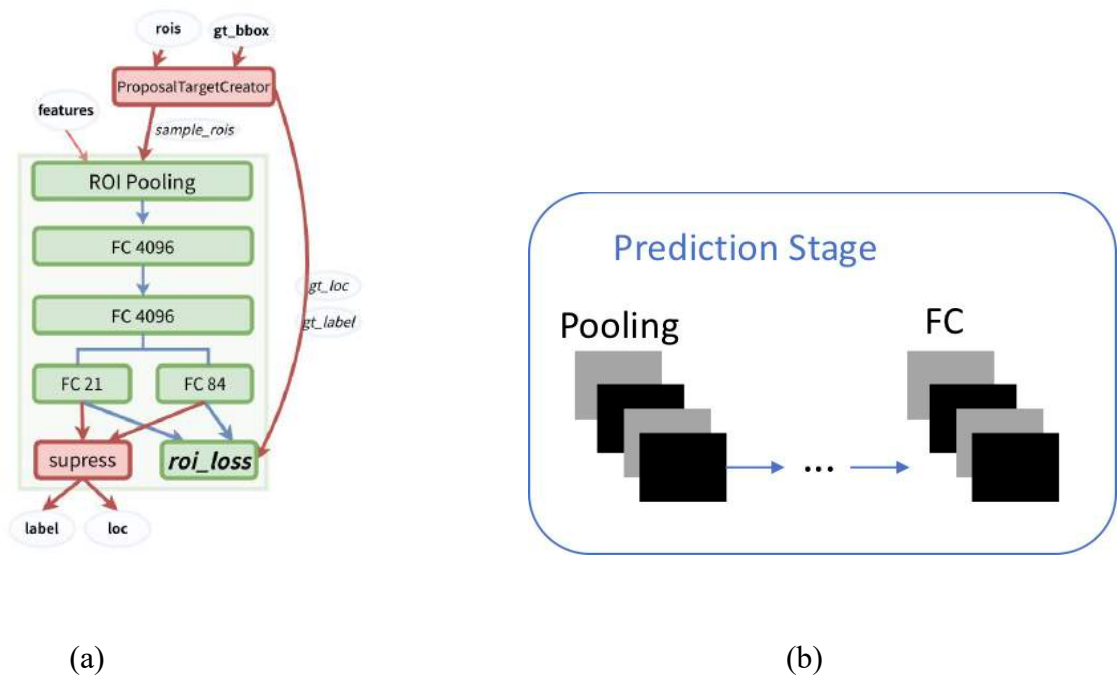

(a)                                                                    (b)

**Figure 3.5**: (a) The whole architecture of Prediction Stage (figure taken from [1]), (b) the high-level architecture of the Prediction Stage.

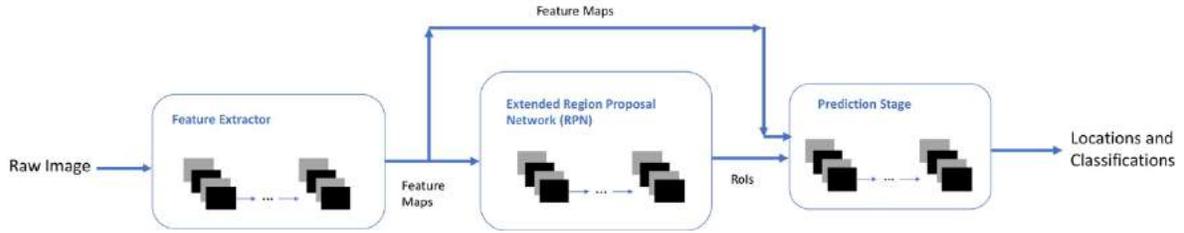

**Figure 3.6**: The illustration of the whole architecture of our advanced model. We incorporate three types of feature extractors: Vgg-16, ResNet-50, and ResNet-101. We incorporate a new technique on ResNet called identity mapping. We propose extended RPN with respect to the original RPN.

### 3.2.2.5 Non-Maximum Suppression (NMS)

For object detection, the state-of-the-art detectors produce multiple overlapping windows close to the region of interest. Non-maximum suppression (NMS) is a key technique that is used to "suppress" multiple overlapping windows, leaving a single bounding box close to the object [32 - 33]. One of the most commonly used approaches for NMS is *Greedy NMS* which applies a greedy iterative method to solve the problem.

*Greedy NMS* iteratively selects the bounding box with the highest score and suppresses other overlapping bounding boxes until no overlapping bounding box exists. Given a bounding-box dataset $B$, a corresponding scores dataset $S$, and an empty dataset $D$, the steps of *Greedy NMS* are organized as follows:

1. Find the highest score $M$ among $S$;
2. Remove the corresponding bounding box $H$ in $B$ with respect to $M$;
3. Add $H$ into dataset $D$;
4. Delete the bounding box in $B$ where the overlapping factor was slightly higher than threshold $N_t$;
5. Iterate steps 1 to 4 until no overlapping bounding box exists.

### 3.2.3 Implementation Details

In this section, we describe our implementation details for both preprocessing and training.

### 3.2.3.1 Preprocessing

The original Faster R-CNN paper suggests that the longer edge of the raw image should be smaller than or equal to 1,000 pixels, and the shorter edge of the raw image should be smaller than or equal to 600 pixels. Thus, we scale KITTI raw image data from size 1242 × 375 to 1000 × 312, which means the scale factor is $\sigma = \frac{1242}{1000} = 1.242$. Meanwhile, we also scale the corresponding ground-truth bounding box coordinates by the same scale factor $\sigma = 1.242$. To make training converges faster, we "normalize" image data, which makes the average value zero. Figure 3.7 illustrates an original image and a normalized image.

### 3.2.3.2 Training

Our model's training procedure is described as follows. We utilized the pre-trained models of Vgg and ResNet, which were pre-trained by He [7] and Simonyan [6] using ImageNet. We froze most of the convolutional layers of Vgg and ResNet. We trained the last three convolutional layers of Vgg, the last four convolutional layers of ResNet, all the layers of the RPN, and all fully-connected layers in the Prediction Stage. We utilized the KITTI 2D object detection dataset as our training dataset. We trained our model to detect cars, pedestrians, cyclist, van, truck, tram, miscellaneous (e.g., trailers, trains), and background.

#### 3.2.3.2.1 RPN training

RPN takes raw image data as input and produces approximately 20,000 anchors as outputs. The training dataset provides ground truth bounding boxes, which are used to compute the Intersection over Union (IoU) with predicted anchors. We assign positive labels to two types of anchors: (1) highest IoU, and (2) IoU is greater than 0.7. Negative labels are assigned to the anchors when IoU is smaller than 0.3. We restrict the total number of positive and negative anchors to 256.

We apply the following loss function for RPN:

$$L = L_{cls}(p_i) + L_{reg}(t_i)$$

in which:

$$L_{cls}(p_i) = \frac{1}{N_{cls}} \sum_i cross\_entropy(p_i, p_i^*)$$

$$L_{reg}(t_i) = \lambda \frac{1}{N_{reg}} \sum_i p_i^* smooth\_L_1(t_i, t_i^*)$$

$$cross\_entropy(p_i, p_i^*) = -p_i \log(p_i^*)$$

$$smooth\_L_1(t_i, t_i^*) = \begin{cases} 0.5(t_i - t_i^*)^2 & if\ |t_i - t_i^*| < 1 \\ |t_i - t_i^*| - 0.5 & otherwise \end{cases}$$

Here, $i$ denotes the index of the anchor, $p_i$ denotes the score of an anchor for a particular type of object, $t_i$ denotes the coordinates of the bounding box, and $N_{cls}$, $N_{reg}$, and $\lambda$ denote normalization parameters. $L_{cls}(p_i)$ is the classification loss function, which is based on the *cross-entropy* loss function. $L_{reg}(t_i)$ is the regression loss function, which is based on the *smooth_$L_1$* loss function. $L$ is the overall loss function.

**3.2.3.2.2 Prediction Stage training**

The Prediction Stage uses a multi-task loss function for training that combines two-stage training into one-stage training. The multi-task loss function is described as follows:

$$L(p, u, t^u, v) = L_{cls}(p, u) + \lambda [u \geq 1] L_{loc}(t^u, v)$$

where $L_{cls}(p, u) = -\log(p_u)$, $L_{loc}(t^u, v) = \sum_i smooth\_L_1(t_i^u, v_i)$,

$$[u \geq 1] = \begin{cases} 1 & u \geq 1 \\ 0 & otherwise \end{cases}$$

Here, $p$ is the probability of each category and background, $u$ and $v$ are the ground-truth category and bounding box coordinates, respectively, and $t^u$ is predicted coordinates of category $u$.

## 3.2.4 Evaluation Details

We evaluate our advanced vehicle detector by adopting the average precision (AP) metric. This section introduces the concepts of confusion matrix, precision, recall, IoU, and average precision (AP). We only evaluate the performance of car category because the KITTI 2D object detection benchmark takes the performance of the car category as the evaluation standard.

## 3.2.4.1 Confusion Matrix

For the object detection task, our goal is to make predictions about object categories and the coordinates of existing objects. For a single ground-truth object, we make "true/false" prediction to denote whether we are able to succeed in detecting the object. However, it is possible that we make a prediction that does not match any ground-truth object. Table 2 illustrates the above concepts as a confusion matrix.

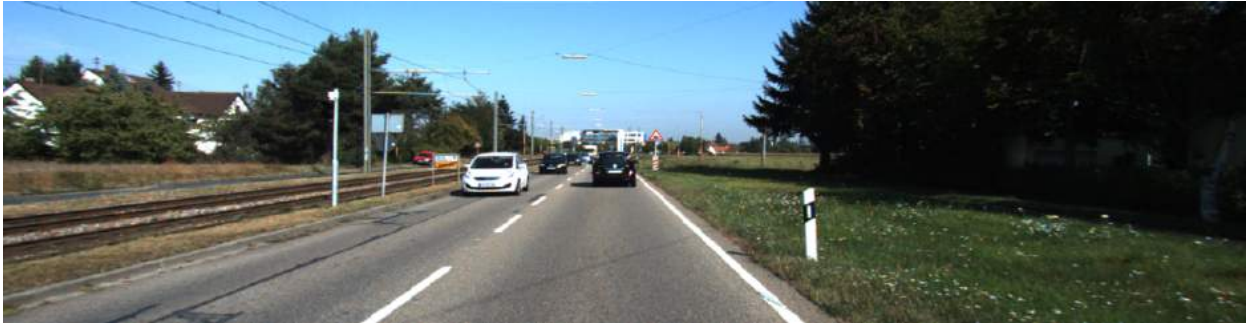

(a)

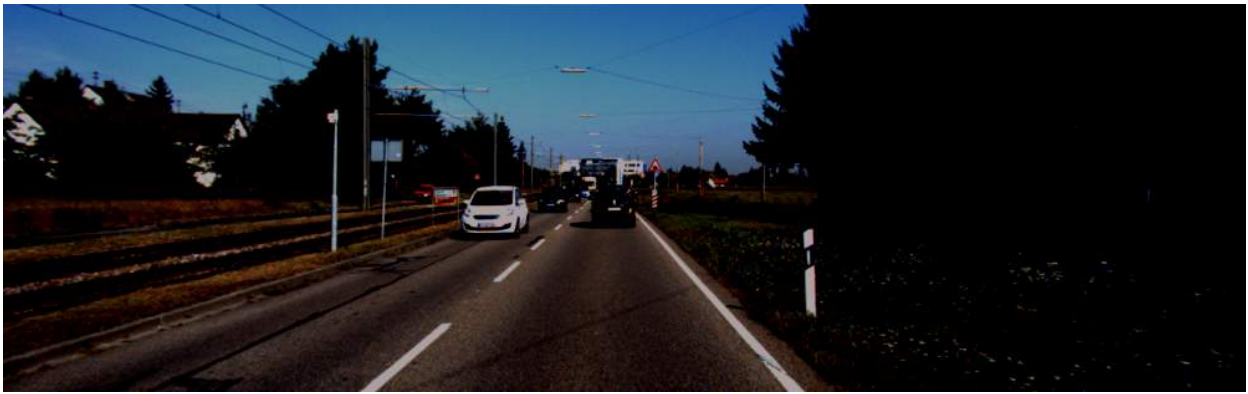

(b)

**Figure 3.7**: (a) The illustration of the original image, (b) the illustration of the "normalized" image.

**Table 2**: Confusion matrix indicating the relationship between ground-truth and predicted label.

|  |  | Ground-Truth |  |
|---|---|---|---|
|  | Total Condition | True | False |
| Predicted | Predicted positive | True Positive (TP) | False Positive (FP) |
|  | Predicted negative | False Negative (FN) | True Negative (TN) |

## 3.2.4.2 Precision and Recall

The definition of precision and recall are as follows:

$$Precision = \frac{TP}{TP + FP}$$

$$Recall = \frac{TP}{TP + FN}$$

Precision measures how accurate your predictions are, indicating the percentage of correct positive predictions. The recall measures how good our detector is at finding all the positives, indicating the percentage of the positive ground-truth objects that our detector finds.

### 3.2.4.3 Intersection over Union (IoU)

The Intersection over Union (IoU) is an evaluation metric in the object detection area. It measures the overlapping percentage between two areas, which is to measure how good our detector is with respect to the ground-truth. We illustrate the definition of IoU in Figure 3.8.

### 3.2.4.4 Average Precision (AP)

Average Precision (AP) is a high-level evaluation metric in object detection, which consists of precision, recall, and IoU. For each ground-truth object, we utilize our detector to make predictions and then judge the correct predictions by comparing IoU with the IoU threshold. We then compute the confusion matrix, precision, and recall. Meanwhile, we also use a confidence score to measure how much confidence we have in our detector. We set the score threshold from 0 to 1 and step size to 0.05, and then we compute the precision and recall pairs. Thus, we plot the PR-curve based on a variety of precision and recall pairs. Average precision is the area under the PR-curve.

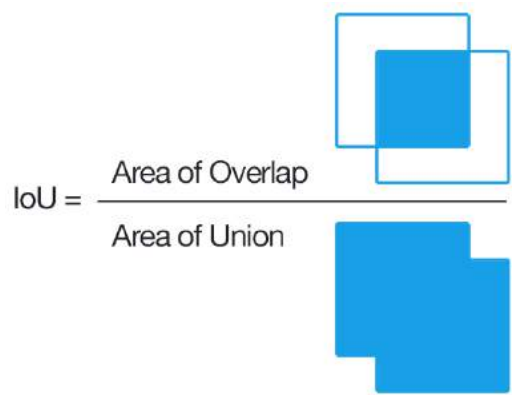

**Figure 3.8**: Illustration and definition of IoU (figure taken from [10]).

# 4
# Experiments

We present our experimental results based on our advanced model, which has been trained and tested on the KITTI 2D object detection dataset. In this section, we share the results of the original Faster R-CNN using feature extractors: Vgg-16, ResNet-50, and ResNet-101. We then present the results of our advanced model to demonstrate the improvements and differences.

## 4.1 Training Results

During our training process, we experimented with several training configurations from $6 \times 10^3$ mini-batches to $3 \times 10^4$ mini-batches, and we concluded that $1.5 \times 10^4$ mini-batches is a suitable choice for our advanced Faster R-CNN training. Figure 4.1 illustrates our training loss. The trend of the training loss starts to saturate at $1.5 \times 10^4$ mini-batches. Thus, we set $1.5 \times 10^4$ mini-batches as the best training configuration based on performance and training time.

## 4.2 Testing Results

We succeeded in detecting the car category of the KITTI vehicle detection dataset using our advanced model. Figure 4.2 illustrates examples results of our advanced model. We also found that the original Faster R-CNN performs poorly in detecting small objects (< 64 pixels). Thus, we propose extended RPN to adapt to small objects. Figure 4.3 illustrates the comparison between the results of the original Faster R-CNN and modified Faster R-CNN.

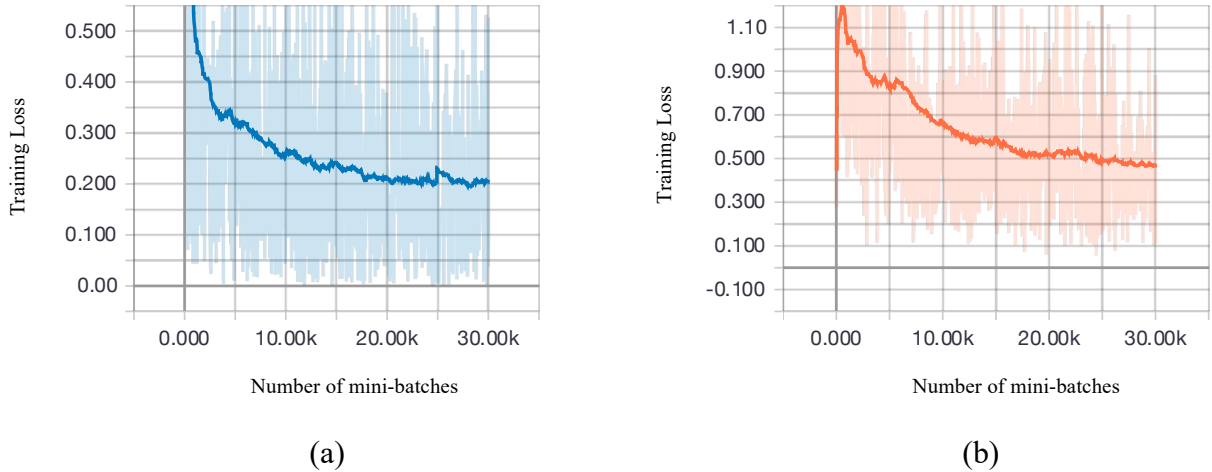

(a) (b)

**Figure 4.1**: (a) The training loss of the whole model, (b) the training loss of the Region Proposal Network (RPN). The bold curves denote the smoothed curves for visualization, and the faint curves denote the original curves.

## 4.3 KITTI 2D Object Detection Benchmark

We evaluated the original Faster R-CNN with the feature extractors Vgg-16, ResNet-50, and ResNet-101 on the KITTI 2D object detection benchmark using the car category. In this section, we present the evaluation results of our advanced model. The KITTI benchmark evaluates results by applying PASCAL VOC criteria, which refers to AP (average precision). The KITTI benchmark also evaluate detection results on three difficulty levels which are listed in Table 3.

We evaluated car category's average precision and draw corresponding PR-curve. We illustrate PR-curves of original Faster R-CNN and our advanced model in Figure 4.4. We also list all the average precision values in Table 4. The whole architecture is trained on NVIDIA GeForce 1080 GPU with 11 GB memory.

**Table 3**: The criteria of three difficulty levels for the KITTI benchmark. We have used the moderate difficulty level.

|  | Min. bounding box height | Max. occlusion level | Max. truncation |
|---|---|---|---|
| Easy | 40 pixels | Fully visible | 15% |
| Moderate | 25 pixels | Partly visible | 30% |
| Hard | 25 pixels | Difficult to see | 50% |

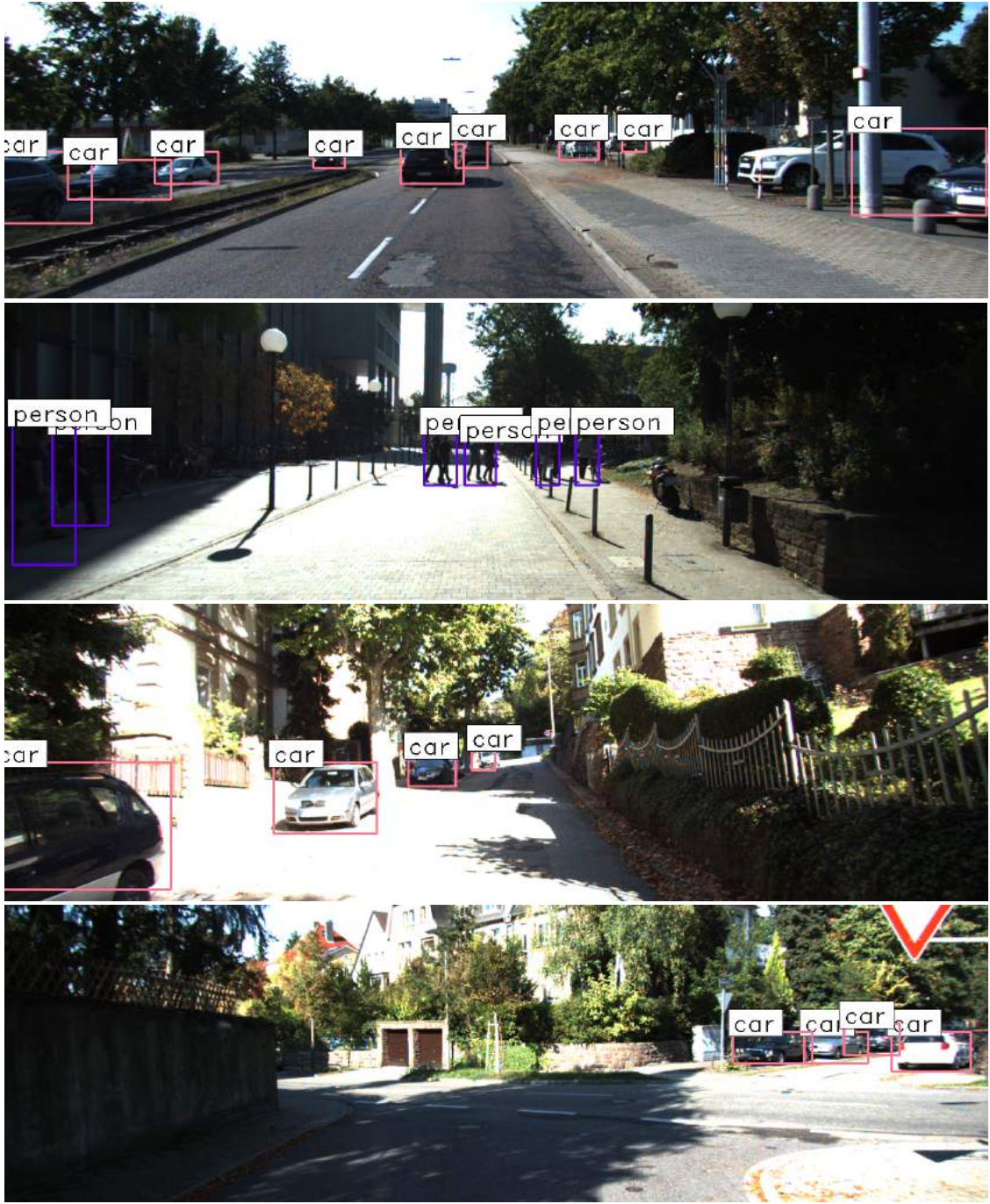

**Figure 4.2**: Testing samples of the KITTI dataset that succeed in detecting cars and persons. The bounding boxes illustrate the location of detected objects. Different colors indicate different categories, and the top texts indicates the category names of each potential object.

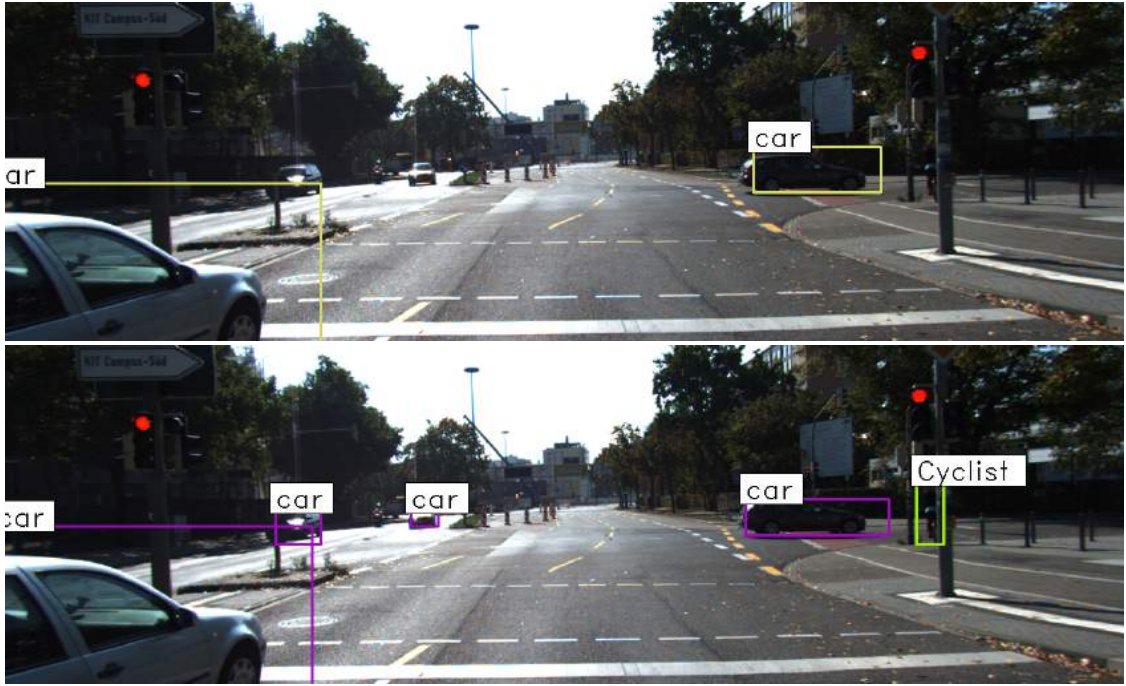

**Figure 4.3**: The **top** image illustrates the testing samples of the original Faster R-CNN with original RPN, and the **bottom** image illustrates the testing samples of our advanced Faster R-CNN with extended RPN. The difference between the two examples is that our model is more sensitive to tiny items.

**Table 4**: The comparisons of our advanced models and original Faster R-CNN models. The **Vgg-16**, **ResNet-50+oRPN**, **ResNet-101+oRPN** indicate the original architectures, and the **ResNet-50+eRPN**, **ResNet-101+oRPN+IM**, **ResNet-101+eRPN+IM** indicate our architectures. The **Vgg-16**, **ResNet-50**, and **ResNet-101** are feature extractors for Faster R-CNN. The **oRPN** and **eRPN** are original RPN and extended RPN which indicate the different anchor sizes of RPN as [128, 256, 512] and [32, 64, 128, 256, 512], respectively. **IM** is the identity mapping technique, which is utilized by ResNet architecture.

| Methods | Moderate AP | Test Time (second/image) |
|---|---|---|
| **Vgg-16** | 57.25% | 0.156 |
| **ResNet-50+oRPN** | 58.63% | 0.203 |
| **ResNet-50+eRPN** | 59.53% | 0.220 |
| **ResNet-101+oRPN** | 65.02% | 0.253 |
| **ResNet-101+oRPN+IM** | 66.29% | 0.250 |
| **ResNet-101+eRPN+IM** | 68.77% | 0.269 |

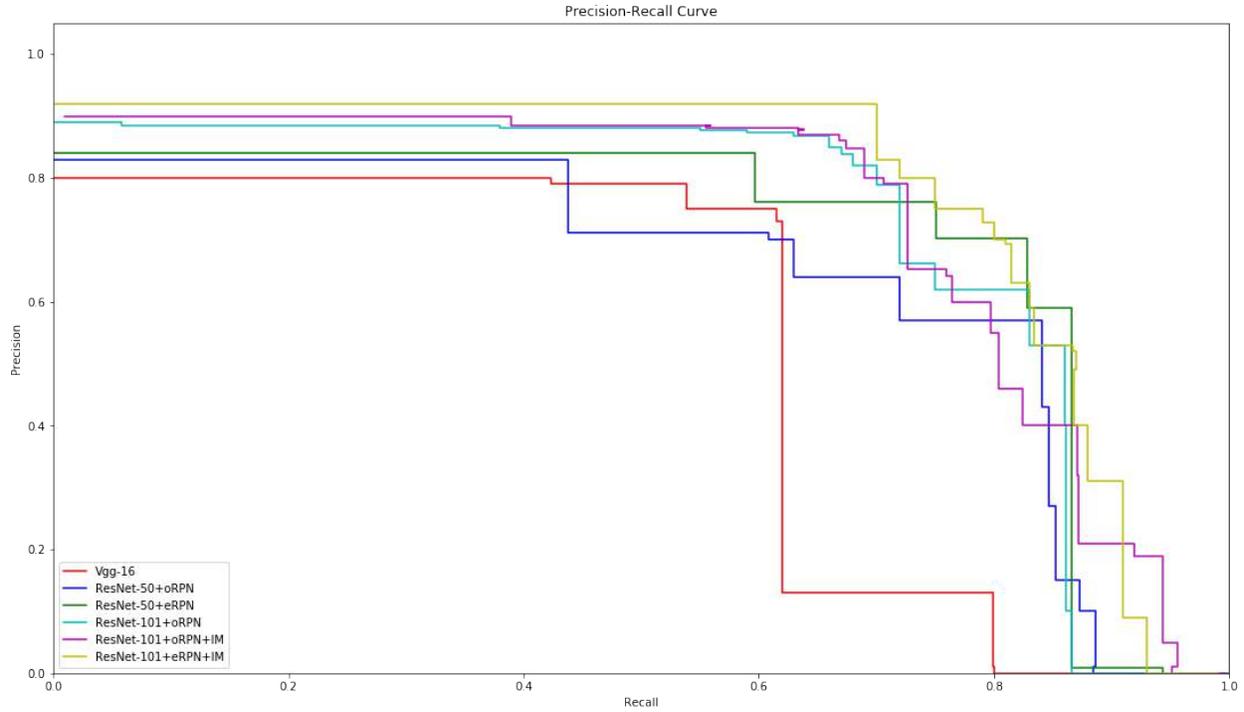

**Figure 4.4**: Precision-Recall curve of car category for average precision (AP) measurement, which is obtained by different models: Vgg-16, ResNet-50+oRPN, ResNet-50+eRPN, ResNet-101+oRPN, ResNet-101+oRPN+IM, and ResNet-101+eRPN+IM. The Vgg-16, ResNet-50+oRPN, ResNet-101+oRPN indicate the original architecture, and the ResNet-50+eRPN, ResNet-101+oRPN+IM, ResNet-101+eRPN+IM indicate our architectures. The Vgg-16, ResNet-50, and ResNet-101 are the feature extractors for Faster R-CNN. The oRPN and eRPN are original RPN and extended RPN which indicate the different anchor sizes of RPN as [128, 256, 512] and [32, 64, 128, 256, 512], respectively. IM is the identity mapping technique, which is utilized by ResNet architecture.

# 5

# Discussion

In this section, we discuss different methods of vehicle detection that have achieved similar performance with respect to our model. We compare and analyze these methods on the KITTI vehicle detection dataset and propose a wide overview of the vehicle detection area. We also discuss the future work based on this research.

## 5.1 Object Detection Methods

In the field of object detection, we reviewed two approaches, which are direct detection, including YOLO [3] and SSD [4], and refined detection, including R-CNN and Faster R-CNN. The designs of these two object detection methods are different, but they achieve similar performance. Refined detection adopts a specialized feature extraction, and then feeds the output of feature extraction into the detector. Direct detection divides the raw image into a 7 × 7 grid such that each grid is a potential bounding box. Additionally, direct detection utilizes the grid cell to predict object locations and then feeds the information into the detector. Thus, the direct detection is faster than the refined detection, but slightly less accurate.

The work most related to our thesis is scale-aware RPN [35]. They utilize a different RPN and feed the output into an XGBoost [36] classifier. Their model achieves 84.81% AP on moderate difficulty level with testing time of 0.9 seconds per image by using a GTX TITAN X GPU with 12 GB memory. Our model achieves 68.77% AP on moderate difficulty level with time of 0.269 seconds per image by using a GTX 1080 with 11 GB memory. A TITAN GPU outperforms a 1080 GPU by 23% in training and testing. In vehicle detection, testing time is as important as accuracy. We think it is necessary to trade-off between testing time and accuracy rather than sacrificing testing time to increase accuracy.

## 5.2 Future Research

Convolutional neural networks perform well and are innovative in the field of image analysis, including object detection, instance segmentation, and object tracking. Combined with large datasets, CNN outperforms traditional machine learning techniques. In vehicle detection, we have proposed a good detection model for two-dimensional vehicle detection, which means we only compute the two-dimensional bounding box of specific objects. People are getting more interested in three-dimensional vehicle detection, which is a great technique for autonomous driving. A popular method in 3D vehicle detection is first computing the 2D bounding box by YOLO or Faster R-CNN, and then determining the dimension and orientation of the objects that are used for the determination of the location of 3D objects.

# 6
# Conclusions

This thesis introduces and improves basic deep learning components that are important in vehicle detection. The implementations and results demonstrate that deep learning techniques have great performance in vehicle detection. The proposed model improves the performance and saves testing time. By incorporating extended RPN, the model is more robust to handle the large variation of vehicle scales.